# Emotion Recognition with Machine Learning Using EEG Signals

Omid Bazgir*
Department of Electrical and Computer Engineering
Texas Tech University
Lubbock, Texas, USA
Email: omid.bazgir@ttu.edu

Zeynab Mohammadi
Department of Electrical and Computer Engineering
University of Tabriz
Tabriz, Iran

Seyed Amir Hassan Habibi
Department of Neurology of Rasool Akram Hospital
Iran University of Medical Sciences
Tehran, Iran

*Abstract*—In this research, an emotion recognition system is developed based on valence/arousal model using electroencephalography (EEG) signals. EEG signals are decomposed into the gamma, beta, alpha and theta frequency bands using discrete wavelet transform (DWT), and spectral features are extracted from each frequency band. Principle component analysis (PCA) is applied to the extracted features by preserving the same dimensionality, as a transform, to make the features mutually uncorrelated. Support vector machine (SVM), K-nearest neighbor (KNN) and artificial neural network (ANN) are used to classify emotional states. The cross-validated SVM with radial basis function (RBF) kernel using extracted features of 10 EEG channels, performs with 91.3% accuracy for arousal and 91.1% accuracy for valence, both in the beta frequency band. Our approach shows better performance compared to existing algorithms applied to the "DEAP" dataset.

## I. INTRODUCTION

Emotion states are associated with wide variety of human feelings, thoughts and behaviors; hence, they affect our ability to act rationally, in cases such as decision-making, perception and human intelligence. Therefore, studies on emotion recognition using emotional signals enhance the brain-computer interface (BCI) systems as an effective subject for clinical applications and human social interactions [1]. Physiological signals are being used to investigate emotional states while considering natural aspects of emotions to elucidate therapeutics for psychological disorders such as autism spectrum disorder (ASD), attention deficit hyperactivity disorder (ADHD) and anxiety disorder [2]. In recent years, developing emotion recognition systems based on EEG signals have become a popular research topic among cognitive scientists.

To design an emotion recognition system using EEG signals, effective feature extraction and optimal classification are the main challenges. EEG signals are non-linear, non-stationary, buried into various sources of noise and are random in nature [3]. Thus, handling and extracting meaningful features from EEG signals plays a crucial role in an effective designing of an emotion recognition system. Extracted features quantify the EEG signals and then are used as attributes of the classifiers. A variety of features have been extracted from the time domain [4], frequency domain [5], and joint time-frequency domain [6] from EEG signals, in intelligent emotion recognition systems. The wavelet transform is able to decompose signals in specific frequency bands with minimum time-resolution loss. To this end, choosing an appropriate mother wavelet is crucial. Murugappan [7] considered four different mother wavelets, namely 'db4', 'db8', 'sym8' and 'coif5' to extract the statistical features, including standard deviation, power and entropy from EEG signal. The KNN classifier was employed to classify five categories of emotions (disgust, happy, surprise, fear, and neutral). The supreme accuracy rate was about 82.87% on 62 channels and 78.57% on 24 channels. Nasehi and Pourghasem [8], applied Gabor function and wavelet transform to extract spectral, spatial and temporal features from four EEG channels. The artificial neural network (ANN) classifier was used for classification of six kinds of emotions (happiness, surprise, anger, fear, disgust and sadness), with 64.78% accuracy. Ishino and Hagiwara extracted mean and variance from power spectra density (PSD), wavelet coefficients of EEG data. Then, a neural network was trained based on the principal components of the features to classify four types of emotion (joy, sorrow, relaxation and anger) with 67.7% classification rate. Mohammedi et al [9], extracted spectral features including energy and entropy of wavelet coefficients from 10 EEG channels. The maximum classification accuracy using KNN was 84% for arousal and 86% for valence. Jie et al [10] employed Kolmogorov-Smirnov (K-S) test to select an appropriate channel for extracting sample entropy as a feature and used it as input of an SVM. The maximum accuracy of Jie's method is 80.43% and 71.16% respectively for arousal and valence. Ali et al [11], combined wavelet energy, wavelet entropy, modified energy and statistical features of EEG signals, using three classifiers including; SVM, KNN, and quadratic discriminant analysis (QDA), to classify emotion states. The overall obtained classification accuracy of Alie's method was 83.8%.

Different theories and principles have been proposed by experts in psychology and cognitive sciences about defining and discriminating emotional states, such as being cognitive or non-cognitive. In this heated debate that continues to go on, there is one claim that states that instead of classifying emotions into boundless common, basic, types of emotions such as; happiness, sadness, anger, fear, joy so on and so forth, emotion can be classified based on valence-arousal model introduced the rooted brain connectivity [12, 13]. The valence-arousal model is a bi-dimensional model which inculdes four emotional states; high arousal high valence (HAHV), high arousal low valence (HALV), low arousal high valence (LAHV) and low arousal low valence (LALV) [1]. Therefore, each of the common emotional states can be modeled and interpreted based on valence-arousal model, figure 1.

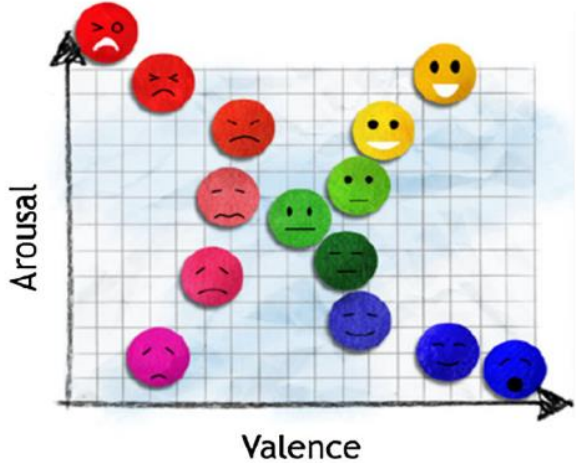

Figure 1. Interpretation of different emotions based on valence-arousal model [12].

In this paper, an emotion recognition system is developed based on the valence-arousal model. We applied wavelet transform on EEG signals, then energy and entropy were extracted from 4 different decomposed frequency bands. PCA was used to make the attributes uncorrelated. SVM, KNN and ANN are utilized for emotional states classification into arousal/valence dimension.

## II. METHODOLOGY

### A. Data Acquisition

In this study, the DEAP database, a database for emotion analysis using physiological signals, labeled based on valence-arousal-dominance emotion model, is used [14]. DEAP dataset includes 32 participants. To stimulate the auditory and visual cortex, 1-min long music videos were played for each participant. 40 music videos were shown to each participants, and seven different modalities were recorded, EEG is used in this study, more information is provided in [14]. The 40 video clips were pre-determined so that their valence/arousal time length would be large enough in valence/arousal scope. Each participant was asked to grade each music video from 1 to 9 for valence, arousal, dominance and liking. Hence, if the grade was greater than 4.5 then the arousal/valence label is high, if the grade is less than 4.5 then the arousal/valence label is low [14]. All the signals were recorded with 512 Hz sampling frequency.

### B. Channel selection

According to Coan et al. [15], positive and negative emotions are respectively associated with left and right frontal brain regions. They have shown that, the brain activity decreases more in the frontal region of the brain as compared to other regions. Therefore, the following channels were selected to investigate in this study: F3-F4, F7-F8, FC1-FC2, FC5- FC6, and FP1- FP2.

### C. Preprocessing

To reduce the electronic amplifier, power line and external interference noise, the average mean reference (AMR) method was utilized. For each selected channel, the mean is calculated and subtracted from every single sample of that channel. To reduce the individual difference effect, all the values were normalized between [0, 1].

### D. Feature extraction

Due to DWT effective multi-resolution capability in analysis of non-stationary signals, we followed our previous work [9] for the feature extraction, in which, DWT was applied on the windowed EEG signals of the selected channels. The EEG signals are windowed due to increasing possibility of the quick detection of the emotional state. Thus, the 4- and 2-seconds temporal windows with 50% overlap were chosen. The EEG signals are decomposed into 5 different bands, including; theta (4-8 Hz), alpha (8-16 Hz), beta (16-32 Hz), gamma (32-64 Hz) and noises (> 64 Hz) via *db4* mother wavelet function. Afterwards, the entropy and energy were extracted from each window of every frequency band.

Entropy is a measurement criterion of the amount of information within the signal. The entropy of signal over a temporal window within a specific frequency band is computed as:

$$ENT_j = -\sum_{k=1}^{N}\bigl(D_j(k)^2\bigr)\log\bigl(D_j(k)^2\bigr) \quad (1)$$

By summing the square of the wavelet coefficients over temporal window, the energy for each frequency band is computed:

$$ENG_j = \sum_{k=1}^{N}\bigl(D_j(k)^2\bigr)\ k = 1,2,\dots,N. \quad (2)$$

Where $j$ is the wavelet decomposition level (frequency band), and $k$ is the number of wavelet coefficients within the $j$ frequency band.

### E. Principal component analysis

PCA is an eigenvector-based statistical mechanism, which employs singular value decomposition, that transforms a set of correlated features into mutually uncorrelated features [16], principal components or PCs. If we define the extracted training features as $X$, then the PCA coefficients ($Z$) can be written as:

$$Z = X\phi \quad (3)$$

Where $\phi$ is the training set underlying basis vector, then the underlying basis vector is applied to the extracted features

of the test set, $\hat{X}$, to obtain principal components of the test set, as:

$$\hat{Z} = \hat{X}\phi \quad (4)$$

The PCA was applied on the stacked extracted features, without any dimensionality reduction, to generate PCs. The PCs are used as the input vectors of the three classifiers addressed in the next section.

*F. Classification*

In this research, kernel SVM, KNN, same as [9], and ANN are used for classification with the eight-fold cross-validation. The goal of SVM, as a parametric classifier, is to formulate a separating hyperplane with application of solving a quadratic optimization problem in the feature space [16]. Kernel SVM finds the optimum hyperplane into a higher dimensional space, that maximizes the generalization capability, where the distance between margins is maximum. The RBF kernel is a function which projects input vectors into a gaussian space, using equation (3). The generalization property makes kernel SVM insensitive to overfitting [17].

$$K_{RBF}(x, x') = \exp[-\sigma\|x - x'\|^2] \quad (5)$$

KNN is a non-parametric instance-based classifier, which classifies an object based on the majority of votes of its neighbors. The votes are being assigned by the K-nearest neighbors' distance to the object.

ANN is a semi-parametric classifier flexible for non-linear classification, which aggregates multilayer logistic regressions [18]. For multilayer feedforward network the nonlinear activation function of the output layer is a sigmoid, equation 4:

$$\phi(x) = \frac{1}{1 + e^{-x}} \quad (6)$$

Which predicts the probability as an output. In the hidden layers rectified linear unit (ReLu) activation function, equation 5, is employed. The ReLu provides sparsity and a reduced likelihood of vanishing gradient, therefore the network convergence is improved and the learning process including backpropagation is faster.

$$\phi(x) = \max(0, x) \quad (7)$$

The implemented ANN incorporates three layers, including two hidden layers with ReLu activation function and the output layer with sigmoid activation function.

Radial basis function (RBF) is implemented as the SVM kernel, with two different scale factors, $\sigma = 2$, $\sigma = 0.1$. KNN is implemented with five different values of nearest neighbors ($3 \leq K \leq 7$), and the optimum classification accuracy is achieved with K=5.

The SVM, ANN and KNN are implemented with eight-fold cross validation to estimate the average accuracy of each classifier. It partitions the data randomly into eight folds, with equivalent size, each fold includes four participants attributes. The learner was trained on seven folds and tested on the remaining fold, the process repeated eight times, each time a different fold is selected for testing.

III. RESULTS

We trained SVM, ANN and KNN classifiers, with extracted features from a pair of channels (F3-F4, F7-F8, FC1-FC2, FC5-FC5, FP1-FP2) of all frequency bands in 2- and 4-seconds temporal window. The RBF kernel of SVM is implemented with $\sigma = 2$. The table 1 shows the cross-validated accuracy of the classifiers with each pair of channels. Optimum accuracy yields with F3-F4 pair of EEG channels, using SVM, which is 90.8 % (sensitivity 88.18%, specificity 89.27%) for arousal and 90.6% (sensitivity 89.9%, specificity 87.7%) for valence. Therefore, to reduce the computational cost, F3-F4 pair of channels can be utilized.

Table 1. Cross-validated accuracy of classifiers from each pair of EEG channels in temporal windows of (a) 4 (b)2 s

| Cross-validated Accuracy (%) | Channels | | | | |
|---|---|---|---|---|---|
| | F3-F4 | F7-F8 | FC1-FC2 | FC5-FC6 | FP1-FP2 |
| **(a)** | | | | | |
| **Arousal-SVM** | **90.8** | 87.9 | 87.2 | 85.1 | 88.1 |
| **Valence-SVM** | **90.6** | 84.9 | 89.8 | 85.5 | 88.5 |
| **Arousal-KNN** | 76.4 | 79 | 75.9 | 77.6 | 71.5 |
| **Valence-KNN** | 79 | 80.4 | 77 | 75.5 | 73.6 |
| **Arousal-ANN** | 82.1 | 83.2 | 71.1 | 77.7 | 80.4 |
| **Valence-ANN** | 84.7 | 83.1 | 73.5 | 74.4 | 78.9 |
| **(b)** | | | | | |
| **Arousal-SVM** | 85.7 | 79.7 | 80.3 | 83.1 | 82.3 |
| **Valence-SVM** | 84.8 | 81.2 | 82.2 | 82.8 | 83.2 |
| **Arousal-KNN** | 68.5 | 70.2 | 65.3 | 66.6 | 64.5 |
| **Valence-KNN** | 69 | 68 | 68.2 | 69.9 | 63.5 |
| **Arousal-ANN** | 74.3 | 76.2 | 73.6 | 70.2 | 73.4 |
| **Valence-ANN** | 71.2 | 79.2 | 69.4 | 68.5 | 74.6 |

Using all the features extracted from all the channels within each frequency band (gamma, beta, alpha, theta), the SVM, ANN and KNN are trained. The cross-validated accuracy of each classifier trained on the principal components of each band is shown in table 2. The trained SVM classifier with beta frequency band features generates the optimum accuracy, 91.3% (sensitivity 89.1%, specificity 88.4%) for arousal and 91.1% (sensitivity 87.3%, specificity 86.8%) for valence.

IV. DISCUSSION AND CONCLUSION

In this research, EEG signals from the DEAP dataset are windowed into 2- and 4-seconds windows with 50% overlap, then, decomposed into 5 frequency bands; gamma, beta, alpha, theta and noise. Spectral features, entropy, and energy of each window within the pre-specified frequency band are extracted. Afterwards, the PCA is applied as a transform preserving the input dimensionality to produce uncorrelated attributes. The ANN, KNN and SVM are trained with attributes from different frequency bands, and different pair of channels, by eight-fold cross validation.

Table 2. Cross-validated accuracy of classifiers from different frequency bands in temporal windows of (a) 4 (b) 2 s

| Cross-validated Accuracy (%) | Frequency Bands | | | |
|---|---|---|---|---|
| | Gamma | Beta | Alpha | Theta |
| **(a)** | | | | |
| **Arousal-SVM** | 89.8 | **91.3** | 90.4 | 89.4 |
| **Valence-SVM** | 89.7 | **91.1** | 90.9 | 89.1 |
| **Arousal-KNN** | 75 | 75.1 | 72.6 | 73.8 |
| **Valence-KNN** | 77 | 78.1 | 78 | 76.1 |
| **Arousal-ANN** | 81.3 | 86.5 | 83.5 | 80.7 |
| **Valence-ANN** | 84.8 | 88.3 | 87.2 | 82.6 |
| **(b)** | | | | |
| **Arousal-SVM** | 81.4 | 89 | 88.68 | 86.6 |
| **Valence-SVM** | 82.43 | 89.34 | 89.72 | 86.1 |
| **Arousal-KNN** | 76 | 73 | 73.7 | 75.6 |
| **Valence-KNN** | 74.3 | 73.2 | 73.8 | 72.4 |
| **Arousal-ANN** | 83.2 | 78.4 | 82.7 | 81.9 |
| **Valence-ANN** | 79.1 | 75.2 | 86.1 | 84.3 |

It is evident from comparing the cross-validated accuracy results shown in table 1 and 2, that the SVM classifier outperforms KNN and ANN. In the previous work [9], the maximum classification accuracy was 86.75% using KNN. In this study, applying PCA made the extracted features uncorrelated. In addition, picking appropriate kernel (RBF), greatly increased the cross-validated accuracy.

Comparing the accuracy rates of tables 1 and 2 shows that, using a 4 seconds window ends up to higher accuracy, therefore, the features are more meaningful. It is also noticeable that emotion status discrimination, based on valence/arousal area, in different frequency bands is higher than the channels. The minimum accuracy in table 2 (a), for valence/arousal SVM is 89.1%, whereas in table 1 (a), for valence/arousal SVM the maximum accuracy is 90.8%. Therefore, it is better to extract features from all the channels in each frequency band, then train a classifier.

Table 3 shows a comparison between our study results and other researchers that have conducted research on the DEAP dataset.

Table 3. Accuracy comparison of different studies on DEAP dataset

| Reference | Year | Number of Channels | Classifier | Accuracy |
|---|---|---|---|---|
| [10] | 2014 | 5 | SVM | 80.4% |
| [11] | 2016 | -- | SVM | 83.8% |
| [9] | 2017 | 10 | KNN | 86.7 % |
| **Our study** | 2018 | 10 | SVM | **91.3%** |

One can increase the accuracy with ensemble learning, using the trained SVM and extracted features in different frequency bands, which makes the algorithm time-consuming or somewhat inappropriate for real-time applications, such as self-driving cars.

For future works, other transforms such as independent component analysis (ICA) or linear discriminant analysis (LDA) could be applied, on the extracted features. Other robust classifiers such as random forest, deep neural network, or recurrent neural network may also lead to higher cross-validated accuracy.

ACKNOWLEDGMENT

The authors are extremely grateful to Dr. Mahmood Amiri of the Kermanshah University of Medical Sciences, and Javad Frounchi from the department of Electrical and Computer Engineering of University of Tabriz for their support and guidance.